# Collaboratively boosting data-driven deep learning and knowledge-guided ontological reasoning for semantic segmentation of remote sensing imagery

*Yansheng Li, Song Ouyang, and Yongjun Zhang*
*School of Remote Sensing and Information Engineering, Wuhan University, Wuhan 430079, China*

**Abstract:** Due to its many potential applications, semantic segmentation of remote sensing (RS) imagery attracts increasing research interest in recent years. As one kind of architecture from the deep learning family, deep semantic segmentation network (DSSN) achieves a certain degree of success on the semantic segmentation task and obviously outperforms the traditional methods based on hand-crafted features. As a classic data-driven technique, DSSN can be trained by an end-to-end mechanism and competent for employing the low-level and mid-level cues (i.e., the discriminative image structure) to understand images, but lacks the high-level inference ability. By contrast, human beings have an excellent inference capacity and can be able to reliably interpret the RS imagery only when human beings master the basic RS domain knowledge. In literature, ontological modeling and reasoning is an ideal way to imitate and employ the domain knowledge of human beings, but is still rarely explored and adopted in the RS domain. To remedy the aforementioned critical limitation of DSSN, this paper proposes a collaboratively boosting framework (CBF) to combine data-driven deep learning module and knowledge-guided ontological reasoning module in an iterative way. The deep learning module adopts the DSSN architecture and takes the integration of the original image and inferred channels as the input of DSSN. In addition, the ontological reasoning module is composed of the intra-taxonomy reasoning function and extra-taxonomy reasoning function. More specifically, the intra-taxonomy reasoning function directly refines the classification result of the deep learning module based on the ontological reasoning rules, and the extra-taxonomy reasoning function aims to generate the inferred channels beyond the current taxonomy set, which are used in the deep learning module, to lift the discriminative performance of confusing objects in the original RS image space. On the one hand, benefiting from the referred channels from the ontological reasoning module, the deep learning module using the integration of the original image and referred channels can achieve better classification performance than only using the original image. On the other hand, a better classification result from the deep learning module further improves the performance of the ontological reasoning module. As a whole, the deep learning module and ontological reasoning module are mutually boosted. Extensive experiments on two publicly open RS datasets such as UCM and ISPRS Potsdam shows that our proposed CBF can outperform the competitive baselines with a large margin.

## 1. Introduction

As a fundamental task in the remote sensing (RS) domain, semantic segmentation of RS imagery [1, 2], which aims to annotate each pixel of the RS imagery with one kind of land-use/land-cover (LULC) type, plays an important role on wide applications such as intelligent agriculture and ecological assessment. Objectively, semantic segmentation of RS imagery is similar to semantic segmentation of natural images in the computer vision domain [3]. However, compared with natural images, the RS imagery often presents complex structures, has much more classification types, and is with arbitrary orientations [4], which bring additional challenges to semantic segmentation of RS imagery [5].

Based on hand-crafted features, shallow classifiers such as support vector machine (SVM) [6], maximum likelihood estimate (MLE) [7] and decision tree (DT) [8], have been widely applied to address semantic segmentation of RS imagery. As a whole, the performance of semantic segmentation methods based on hand-crafted features and shallow classifiers is still very limited. Along with the rapid development of deep networks such as deep detection networks [9, 10], deep recognition networks [11] and deep hashing networks [12,13], deep semantic segmentation network (DSSN) has been fully exploited on semantic segmentation of RS imagery [14][15][16][17]. As well known, DSSN is one kind of data-driven technique, and its obvious superiority is that the hyperparameters of DSSN can be learned via an end-to-end manner. Due to the black-box learning characteristic of deep learning, the interpretability and reliability of DSSN is still very limited [18]. Hence, how to reinforce DSSN to further improve the semantic segmentation performance deserves much more exploration.

DSSN can be competent for employing the low-level and mid-level cues to understand images, but lacks the high-level inference ability. By contrast, human beings have an excellent inference capacity and can be able to reliably interpret the RS imagery only when human beings master the basic RS domain knowledge. The reason why RS experts can quickly and accurately make classification is because they have the necessary prior knowledge of geoscience and the ability to obtain new knowledge through knowledge reasoning. Although data-driven learning methods are the mainstream, knowledge-driven learning methods are still regarded as one of the most important research directions by RS community [19]. Since the prior knowledge of geoscience is basic common sense and the reasoning results are foreseeable, expressing the prior knowledge through formal language and forming inference ability can establish a knowledge-driven interpretable method of semantic segmentation of RS imagery. However, not all prior knowledge of geoscience can be easily modeled formally, which greatly limits the practical performance of knowledge-based methods. In this context, the future of RS science should be supported by knowledge representation technologies such as ontology. Ontology [20], as a formal expression of concepts and their relationships in a specific domain, has strong knowledge representation capabilities, inference capabilities based on cognitive semantics, and the ability to share knowledge [21]. By introducing ontology for semantic segmentation of RS imagery, knowledge reasoning can be performed with the ontology which expresses the symbolized prior knowledge of RS experts. ontology-based semantic reasoning can fully mine the rich semantic information between objects with expert knowledge. However, the performance of ontological reasoning for semantic segmentation of RS imagery is still very limited, so the mechanism of ontological reasoning still needs more research.

Therefore, how to effectively combine data-driven learning methods with knowledge-driven learning methods is a promising way to pursue the explainable artificial intelligence (AI) [22]. In addition, coupling data-driven deep learning and knowledge-guided ontological reasoning is a rational way to achieve the truly intelligent interpretation of RS imagery. Intuitively, the combination of ontological reasoning and deep learning can make full use of the advantages of knowledge-driven and data-driven methods. On the one hand, it helps to directly correct the misclassification and improve the interpretability of the classification results. On the other hand, additional information, which is generated by ontological reasoning, benefits enhancing the deep network's anti-interference ability. Thereby, ontological reasoning effectively solves the problem of data-driven method's poor anti-interference ability and lack of interpretation.

Based on the above analysis, we propose a collaboratively boosting framework (CBF) to combine bottom-up data-driven deep learning and top-down knowledge-guided ontological reasoning for semantic segmentation of high-resolution RS imagery, which realizes the interaction between reasoner and classifiers. It not only effectively learns the low-level and mid-level cues which are difficult to accurately express by DSSN, but also combines ontological reasoning to make the results interpretable and credible. Intra-taxonomy ontological reasoning rules and extra-taxonomy ontological reasoning rules are designed for reasoner. The former is used to directly correct the misclassification of the DSSN-based classifier, and the latter is used to extract the estimation of shadow and elevation from the corrected results as additional information to enhance the anti-interference capability of the DSSN-based classifier. The DSSN autonomously learns low-level and mid-level features from the data. Ontology reasoner uses high-level domain knowledge to guide the interpretation including correcting misclassification directly and extracting additional information to assist the DSSN indirectly. The whole process forms a closed loop and iterates continuously until the accuracy of classification converges. Our proposed CBF is tested on two publicly open RS datasets such as UCM and ISPRS Potsdam. It not only improves the interpretability and reliability of the classification, but also further promotes the classification accuracy. As a whole, the main contributions of this paper are summarized as follows:

- This paper proposes a CBF which can mutually reinforce data-driven deep learning and knowledge-guided ontological reasoning in an iterative way. It is worth noting that CBF is a general idea and more variants can be designed based on the specific task requirements.
- We present a new unified ontological reasoning approach which includes the intra-taxonomy reasoning function and extra-taxonomy reasoning function. The intra-taxonomy reasoning function directly refines the classification result of the deep learning module. In addition, the extra-taxonomy reasoning function benefits improving the deep learning module from the input perspective.

The rest of this article is organized as follows. Section 2 describes the related work. The proposed method is detailed in Section 3. Section 4 analyzes and discusses the experimental results. Finally, Section 5 summarizes the work of this paper and points out some potential research directions.

## 2. Related Work

In recent years, deep learning has been widely used in semantic segmentation of high-resolution RS imagery. Basaeed et al. [23] used a convolutional neural network (CNN) to perform multi-scale analysis on each channel, which performed fusion and morphological operations on the obtained boundary confidence map to obtain a hierarchical segmentation map. Langkvist et al. [24, 25] applied deep DSSN to multispectral imagery (MSI) to achieve fast and accurate pixel-by-pixel classification. Audebert et al. [26] trained variants of the SegNet structure and introduced multi-core convolutional layers to quickly aggregate predictions on multiple scales. Maggiori et al. [27] designed a CNN which was able to learn how to combine features at different resolutions to integrate local and global information in an efficient and flexible manner. Kampffmeyer et al. [28] proposed a novel deep CNN, which is mainly used for land cover mapping in urban areas in RS imagery. It detected small objects effectively while achieving high overall accuracy. In general, all of these existing deep learning-based methods follow a data-driven learning mechanism, and still cannot make full use of the high-level knowledge of domain experts. As a consequence, these methods are susceptible to noise attack.

On another research avenue, ontology-based knowledge models have great advantages in expressing and applying knowledge. By solving the major limitations of deep learning methods used in RS science in knowledge cognition, ontology-based knowledge models have great potential for long-term advancement of RS [29]. Codescu et al. [30] applied OSM ontology to geographic information system (GIS), but the application of the ontology is more limited. Sarker et al. [21] proposed a system for interpreting the output of a classifier based on a knowledge model, which gave an explanation of the classification, but the classification accuracy was poor. Samuel et al. [31] used the ontology to classify the Landsat images based on explicit spectral rules, but use spectral information only, without take other information such as the shape, texture, and spatial relationship of objects into account. Khitem et al. [32] proposed a method for semantic labeling of RS imagery based on regional adjacency maps. This method uses the spatial and spectral attributes of objects in ontology to complete the labeling. Gui et al. [33] used ontology to extract buildings from TerraSAR-X RS imagery. Geographic object-based image analysis (GEOBIA) uses the knowledge of domain experts to extract information such as the shape, texture, and spatial relationships of objects to complete the classification of objects. Gu et al. [34] proposed an ontology-based semantic classification method for high-resolution RS imagery, which aims to make full use of the advantages of GEOBIA and ontology. Ontological reasoning in above methods enhances the interpretability and credibility of the classification, but its classification performance is still very limited compared with deep learning-based methods as modeling all of the domain knowledge is still an open problem.

Apparently, combining deep learning and ontological reasoning is a promising way to coordinate data-driven and knowledge-driven methods [35]. As a first attempt towards this direction, Alirezaie et al. [36] proposed a method to combine the ontology reasoner and the DSSN-based classifier. The research showed that, as an additional input to the DSSN, extra information (e.g., shadows and elevations) can effectively improve the accuracy of the classification. However, this method presents two limitations. The first one is that the ontological reasoning is only indirectly involved in the classification process and does not directly guide the classification. The other one is that the extra information obtained by direct reasoning on the output of the misclassification is not accurate. Therefore, how to make full use of the combination merits requires further exploration.

## 3. Methodology

As visually shown in Fig. 1, our proposed CBF includes two main modules (i.e., the DSSN-based classification module and ontological reasoning module). More specifically, ontological reasoning is composed of intra-taxonomy reasoning and extra-reasoning. Overall, our proposed method is trained via an iterative manner. Fig. 1 shows the workflow of the training phase. In each iteration of the training phrase, DSSN is firstly trained. Then, the intra-taxonomy ontology reasoner directly corrects the misclassification from the output of the DSSN according to the inference rules. The extra-taxonomy ontology reasoner extracts additional information such as more accurate shadow and elevation information based on the corrected classification. Finally, the additional information is used as the additional channel for training DSSN in the next iteration.

It is noted that, in the testing phase, only the trained DSSN and the intra-taxonomy reasoning are adopted. Given a fixed iteration, the output of the trained DSSN is denoted as the result of Stage I and the refined result by the intra-taxonomy reasoning function is denoted by the result of Stage II.

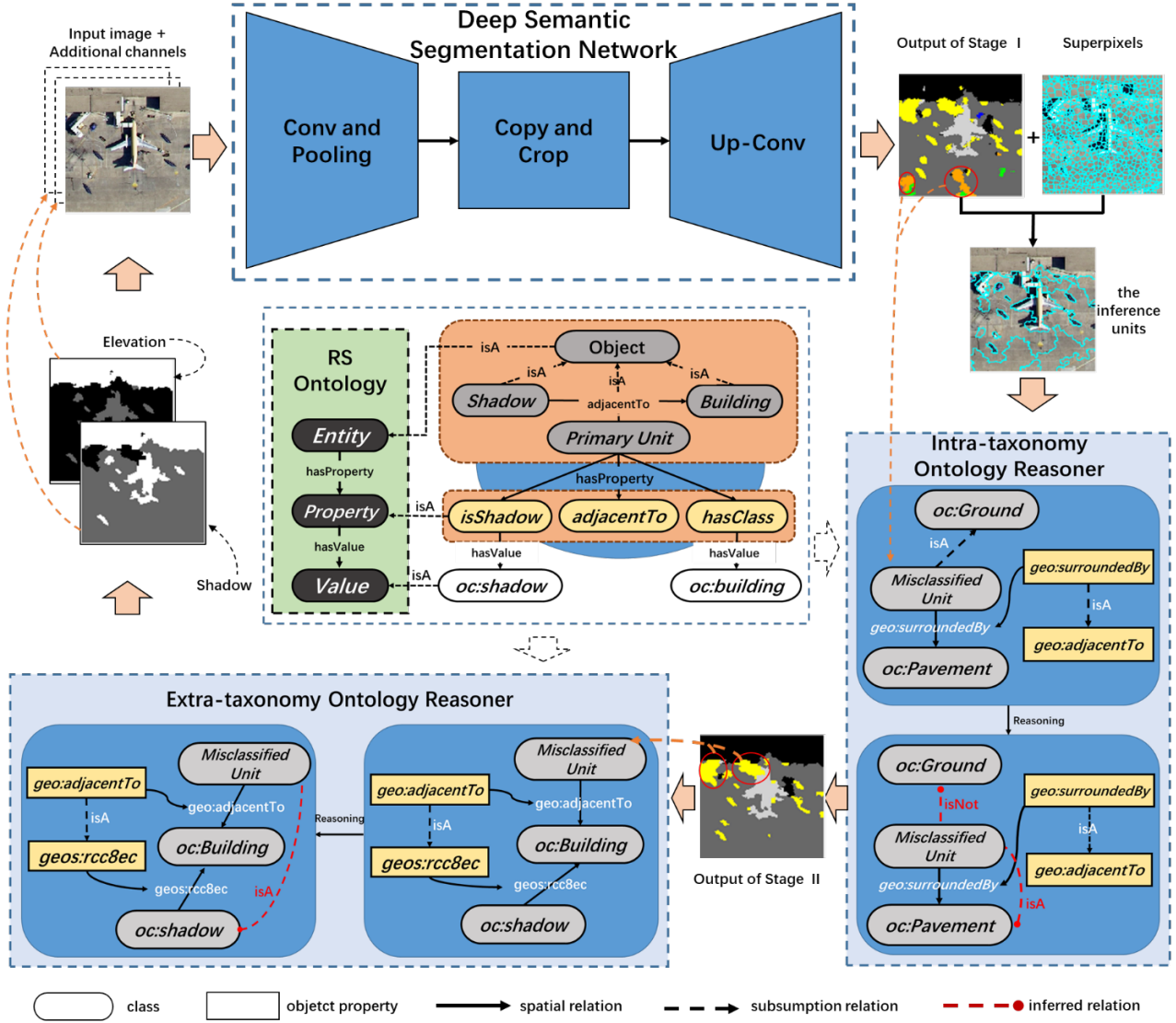

Fig. 1 The workflow of the proposed CBF.

**3.1. Learning deep semantic segmentation network (DSSN)**

In this paper, U-Net [37], which is a classic image semantic segmentation network based on a full convolutional neural network, is adopted as the backbone of DSSN.

The input image with the inferenced additional channels and its corresponding label are used to train the DSSN. Assume that the original data of the RS imagery is *I* and its corresponding additional channels are *E* (*E* is set to 0 in the first iteration step), and let θ stands for the hyperparameter of the DSSN. The forward prediction probability that a pixel with the image coordinate (i,j) belongs to the *c-th* class is denoted by $p_c(i,j)$, as shown in Eq. (3).

$$p_c(i,j) = \varphi\big((I, E), \theta\big) \qquad (3)$$

where φ represents the hierarchical mapping function of DSSN.

Based on the optimization objective function $J$ via the traditional cross-entropy loss, as in Eq. (4), DSSN is updated by the backward propagation algorithm.

$$J = -\sum_i \sum_j \left( \sum_{c=1}^{n} y_c(i,j) \log(p_c(i,j)) \right) \tag{4}$$

where $n$ is the number of class, $y_c(i,j)$ is the forward prediction result of the DSSN. If the class is the same as the sample, $y_c(i,j)$ is 1, otherwise it is 0.

### 3.2. Generating inference units

First, in the first iteration, establish ontology for the interpretation of remote sensing (RSOntology), intra-taxonomy ontological reasoning rules and extra-taxonomy ontological reasoning rules, and add the rules to the rule base of ontology; Then use the sample image $I$ and the additional channels $E$ to train the DSSN and classify the RS imagery (the additional channels are set to zero at the first iteration step, and the subsequent iteration steps can be adaptively generated), so as to obtain the category $C$ of each pixel and its classification confidence $F$. Next, superpixel segmentation is performed on the image $I$ to obtain a superpixel set $G$ with $K$ superpixels, as shown in Eq. (1).

$$G = \{S_1, S_2, \ldots, S_k | S_i = Segment(I), 1 \leq i < K\} \tag{1}$$

where $S_i$ denotes the $i-th$ superpixel.

The class of the most pixels among all the pixels in each superpixel region is taken as the category of the superpixel. Furthermore, the superpixels are clustered according to the class and the spatial neighboring relationship. After the clustering-based aggregation, the superpixels are taken as the inference unit $S'$, as shown in Eq. (2).

$$S' = \{S_i | C_i = C, S_i \text{ Adjacent to } S, 1 \leq i < K\} \tag{2}$$

Then calculate the average confidence of all pixels in the inference unit $S'$ as the classification confidence of the inference unit $S'$. The classification confidence $F$ can be used as a basis for judging whether the classification is correct or not. An inference unit with low confidence (i.e., $F < F_t$) is taken as misclassification unit where $F_t$ is an empirical threshold. Then calculate the spatial relationship (such as adjacency, orientation, inclusion, etc.) between adjacent inference units, and combine the attribute of the inference unit and the intra-taxonomy ontological reasoning rules to perform ontological reasoning, so as to correct the misclassification to get the map of corrected classification; Next, perform the ontological reasoning on the map of corrected classification according to the extra-taxonomy ontological reasoning rules to extract the information of shadow and relative elevation; Finally, the information as additional channels together with the corresponding raw image are sent to DSSN for the next iteration until the optimization objective function converges.

**3.3. Constructing ontology of remote sensing (RSOntology)**

As the interpretation of RS imagery, RSOntology is used to describe the attributes of objects and the relationships between objects. In order to establish the ontology, the ontology web Language (OWL) which is a formal language, is used to express the RSOntology.

As shown in Fig. 2, the root class in ontology is $oc: GeoObject$ which mainly includes the class of the inference unit ($oc: Segment$), vegetation ($oc: Vegetation$), bare land ($oc: Ground$), road ($oc: Pavement$), building ($oc: Building$), water ($oc: Water$), Airplane ($oc: Airplane$), vehicle ($oc: Car$), ship ($oc: Ship$), etc. $oc: Segment$ contains the class of classified unit ($oc: ClassifiedSegment$) and the class of misclassified unit ($oc: MisClassifiedSegment$). $oc: geoClass$ is the class of object. The core attributes are defined in ontology. $oc: isA$ represents the subordinate attribute. The attribute of spatial relationship is composed of adjacency ($geo: adjacentTo$), surround ($geo: surroundedBy$) and direction ($geo: hasDirectionOf$). $geo: MaxClass$ is a statistical property, which represents the value of the most class.

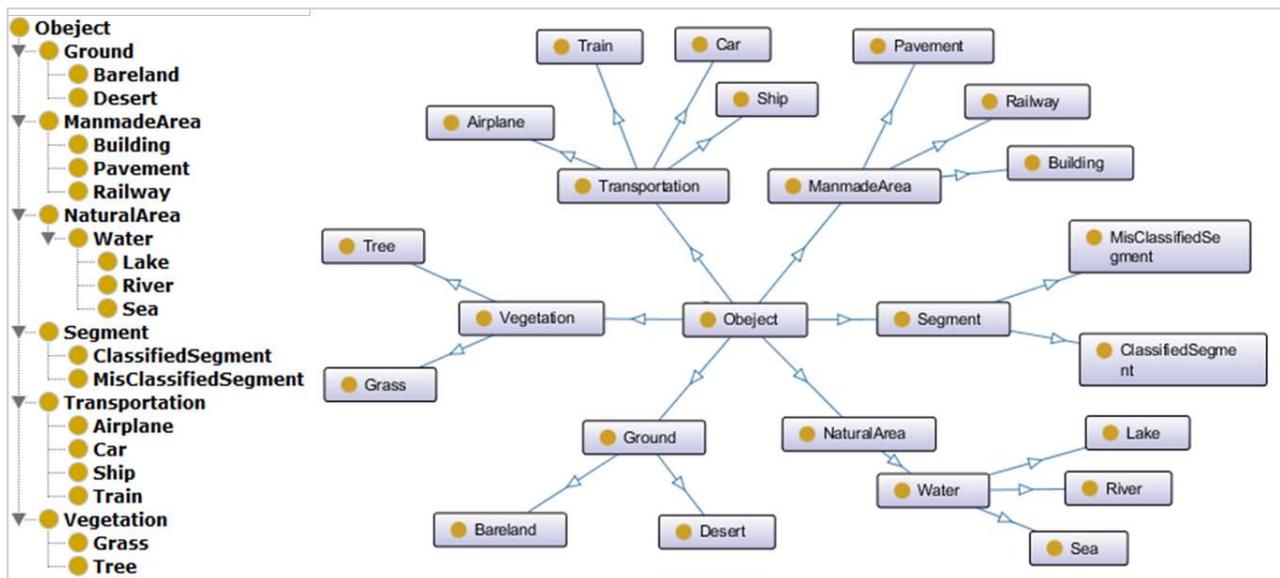

Fig. 2. The visual illustration of RSOntology.

**3.4. Ontological reasoning**

Ontological reasoning in this method includes intra-taxonomy ontological reasoning and extra-taxonomy ontological reasoning. The former directly corrects the misclassification according to inference rules, and the latter extracts the additional information of shadow and relative elevation to assist the classification indirectly. The superpixel $S'$ after superpixel clustering in Section 2 is used as the basic inference unit. It uses Description Logic (DL) [27] to symbolize the ontological reasoning rules. $oe\!:\!entity$ and $oe\!:\!entity1$ are instances of the inference unit that is correctly classified, satisfying Eq. (5) and Eq. (6). $oe\!:\!misEntity$ is the instance of the inference units that is misclassified, satisfying equations Eq. (7).

$$oc\!:\!ClassifiedSegment(oe\!:\!entity) \tag{5}$$

$$oc\!:\!ClassifiedSegment(oe\!:\!entity1) \tag{6}$$

$$oc\!:\!MisClassifiedSegment(oe\!:\!misEntity) \tag{7}$$

### 3.4.1 Intra-taxonomy ontological reasoning

The rules of intra-taxonomy ontological reasoning are symbolized by using DL [38], including two types of rules, as shown in Table 1.

One type of rule is used to eliminate the hole phenomenon caused by misclassification, that is, to correct the classification of hole area as the class of surrounding objects, including rules 1 to 6. For example, if a building which is misclassification in the results is surrounded by water, the true classification of the building area should be water.

Table 1. The rules of intra-taxonomy ontological reasoning.

| Num | Description | Expression based on DL |
|---|---|---|
| Rule 1 | If an inference unit, misclassified as vegetation, is surrounded by ground, pavements, buildings or water, its class should be corrected to the category of object around it. | $oc\!:\!Vegetation(oe\!:\!misEntity)$, $oc\!:\!geoClass \sqsubseteq oc\!:\!Ground \sqcup oc\!:\!geoClass \sqsubseteq oc\!:\!Pavement$ $\sqcup oc\!:\!geoClass \sqsubseteq oc\!:\!Building \sqcup oc\!:\!geoClass \sqsubseteq oc\!:\!Water$, $\forall oe\!:\!entity \in (\forall geo\!:\!adjacentTo.oe\!:\!misEntity)$ $oc\!:\!geoClass(oe\!:\!entity)$, $\Rightarrow oc\!:\!geoClass(oe\!:\!misEntity)$ |
| Rule 2 | If an inference unit, misclassified as ground, is surrounded by pavements, buildings or water, its class should be corrected to the category of object around it. | $oc\!:\!Ground(oe\!:\!misEntity)$, $oc\!:\!geoClass \sqsubseteq oc\!:\!Pavement \sqcup oc\!:\!geoClass \sqsubseteq oc\!:\!Building$ $\sqcup oc\!:\!geoClass \sqsubseteq oc\!:\!Water$, $\forall oe\!:\!entity \in (\forall geo\!:\!adjacentTo.oe\!:\!misEntity)$ $oc\!:\!geoClass(oe\!:\!entity)$, $\Rightarrow oc\!:\!geoClass(oe\!:\!misEntity)$ |
| Rule 3 | If an inference unit, misclassified as building, is surrounded by ground or water, its class should be corrected to the category of object around it. | $oc\!:\!Building(oe\!:\!misEntity)$, $oc\!:\!geoClass \sqsubseteq oc\!:\!Ground \sqcup oc\!:\!geoClass \sqsubseteq oc\!:\!Water$, $\forall oe\!:\!entity \in (\forall geo\!:\!adjacentTo.oe\!:\!misEntity)$ $oc\!:\!geoClass(oe\!:\!entity)$, $\Rightarrow oc\!:\!geoClass(oe\!:\!misEntity)$ |

| Rule 4 | If an inference unit, misclassified as water, is surrounded by vegetation, buildings or pavements, its class should be corrected to the category of object around it. | $oc: Water(oe: misEntity),$ <br> $oc: geoClass \sqsubseteq oc: Vegetation \sqcup oc: geoClass$ <br> $\sqsubseteq oc: Building$ <br> $\sqcup oc: geoClass \sqsubseteq oc: Pavement$ <br> $\forall oe: entity \in (\forall geo: adjacentTo.oe: misEntity)$ <br> $oc: geoClass(oe: entity),$ <br> $\Rightarrow oc: geoClass(oe: misEntity)$ |
|---|---|---|
| Rule 5 | If an inference unit, misclassified as airplane, is surrounded by vegetation, buildings or water, its class should be corrected to the category of object around it. | $oc: Airplane(oe: misEntity),$ <br> $oc: geoClass \sqsubseteq Vegetation \sqcup oc: geoClass \sqsubseteq oc: Building$ <br> $\sqcup oc: geoClass \sqsubseteq oc: Water,$ <br> $\forall oe: entity \in (\forall geo: adjacentTo.oe: misEntity)$ <br> $oc: geoClass(oe: entity),$ <br> $\Rightarrow oc: geoClass(oe: misEntity)$ |
| Rule 6 | If an inference unit, misclassified as car, is surrounded by vegetation or water, its class should be corrected to the category of object around it. | $oc: Car(oe: misEntity),$ <br> $oc: geoClass \sqsubseteq oc: Vegetation \sqcup oc: geoClass \sqsubseteq oc: Water$ <br> $\forall oe: entity \in (\forall geo: adjacentTo.oe: misEntity)$ <br> $oc: geoClass(oe: entity),$ <br> $\Rightarrow oc: geoClass(oe: misEntity)$ |
| Rule 7 | If an inference unit is misclassified as airplane, with none of the correctly classified objects which are pavement adjacent to it, its class should be the category with the most correctly classified objects in its neighborhood. | $oc: Airplane(oe: misEntity),$ <br> $\forall oe: entity \in (\forall geo: adjacentTo.oe: misEntity)$ <br> $oe: entity \in \neg oc: Pavement,$ <br> $\Rightarrow oe: misEntity \in geo: MaxClass(oe: entity)$ |
| Rule 8 | If an inference unit is misclassified as car, with none of the correctly classified objects which are pavement adjacent to it, its class should be the category with the most correctly classified objects in its neighborhood. | $oc: Car(oe: misEntity),$ <br> $\forall oe: entity \in (\forall geo: adjacentTo.oe: misEntity)$ <br> $oe: entity \in \neg oc: Pavement,$ <br> $\Rightarrow oe: misEntity \in geo: MaxClass(oe: entity)$ |
| Rule 9 | If an inference unit is misclassified as ship, with none of the correctly classified objects which are water adjacent to it, its class should be the category with the most correctly classified objects in its neighborhood. | $oc: Ship(oe: misEntity),$ <br> $\forall oe: entity \in (\forall geo: adjacentTo.oe: misEntity)$ <br> $oe: entity \in \neg oc: Water,$ <br> $\Rightarrow oe: misEntity \in geo: MaxClass(oe: entity)$ |

The other type of rules is used to correct misclassifications due to inconsistencies in spatial relationships. The true classification is most likely to be the class of the most surrounding category, including rules 7-9. For example, the misclassified vehicle is not adjacent to the road, so its class is more likely the category with the most correctly classified objects in its neighborhood.

### 3.4.2 Extra-taxonomy ontological reasoning

The rules of extra-taxonomy ontological reasoning are symbolized by using DL, including two types of rules, as shown in Table 2.

One type of rules is used to extract shadow, including rules 1 to 4. For example, If an objects that is misclassified as pavements, ground, water or cars is adjacent to correctly classified buildings, there is a shadow in the corresponding areas. In the extra channel, pixels are assigned with 1 (shadow), 0 (uncertain) and -1 (not shadow).

Table 2. The rules of Extra-taxonomy ontological reasoning

| Num | Description | Expression based on DL |
|---|---|---|
| Rule 1 | The misclassification is pavement, ground, water or car. If there is a correctly classified building in its neighborhood, there is shadow in the area. | $oc{:}geoClass \sqsubseteq oc{:}Pavement \sqcup oc{:}Ground \sqcup oc{:}Water \sqcup ec{:}Car$, $oc{:}geoClass(oe{:}misEntity)$, $\exists oe{:}entitiy \in (\forall geo{:}adjacentTo.oe{:}misEntity)$ $oc{:}Building(oe{:}entitiy)$, $\Rightarrow oc{:}Shadow(oe{:}misEntity)$ |
| Rule 2 | The misclassification is car, ship, vegetation or airplane. If there are no correctly classified buildings in its neighborhood, there is no shadow in the area. | $oc{:}geoClass \sqsubseteq oc{:}Vegetation \sqcup oc{:}Car \sqcup oc{:}Ship \sqcup oc{:}Airplane$, $oc{:}geoClass(oe{:}entity)$, $\exists oe{:}entitiy \in (\forall geo{:}adjacentTo.oe{:}misEntity)$ $oe{:}entitiy \in \neg oc{:}Building$, $\Rightarrow oc{:}NonShadow(oe{:}entity)$ |
| Rule 3 | The correct classification is ground. If there are no correctly classified buildings and vegetation in its neighborhood, there is no shadow in the area. | $oc{:}Ground(oe{:}entity)$, $\exists oe{:}entitiy1 \in (\forall geo{:}adjacentTo.oe{:}entity)$ $oe{:}entitiy1 \in (\neg oc{:}Building \sqcap \neg oc{:}Vegetation)$, $\Rightarrow oc{:}NonShadow(oe{:}entity)$ |
| Rule 4 | If the correct classification is building, there is no shadow in the area. | $oc{:}Building(oe{:}entity)$, $\Rightarrow oc{:}NonShadow(oe{:}entity)$ |
| Rule 5 | if an object is correctly classified as vegetation, ground, pavements or water, it has low elevation. | $oc{:}geoClass \sqsubseteq oc{:}Vegetation \sqcup oc{:}Ground \sqcup oc{:}Pavement \sqcup oc{:}Water$, $oc{:}geoClass(oe{:}entity)$, $\Rightarrow geo{:}hasLowElevation(oe{:}entity)$ |
| Rule 6 | if an object is correctly classified as airplane, car or ship, it has medium elevation. | $oc{:}geoClass \sqsubseteq oc{:}Airplane \sqcup oc{:}Car \sqcup oc{:}Ship$, $oc{:}geoClass(oe{:}entity)$, $\Rightarrow geo{:}hasMediumElevation(oe{:}entity)$ |
| Rule 7 | if an object is correctly classified as building, it has high elevation. | $oc{:}Buliding(oe{:}entity)$, $\Rightarrow geo{:}hasHighElevation(oe{:}entity)$ |

Another type of rule is used to extract the relative elevation, including rules 5-7. For example, if an object is correctly classified as building, it has high elevation. In the extra channel, pixels are assigned with 2 (high elevation), 1 (medium elevation) and 0 (low elevation).

## 4. Experimental results and discussion

### 4.1. Evaluation datasets

The experiments use the publicly open UCM dataset [39] and the publicly open Potsdam dataset [40].

The UCM dataset contains 21 classes. each class has 100 images with 256x256 in size. The ground resolution of images is 0.3m. The sample set uses the densely labeled DLRSD dataset [41] from UCM dataset, which contains 17 classes. In order to reduce the similarity between the classes, this paper merges the 17 classes into 8 classes, which are Vegetable (trees, grass), Ground (bare soil, sand, chaparral), Pavement (pavement, dock), Building ( building, mobile home, tank), Water (water, sea), Airplane (airplane), Car (car) and Ship (ship), and removes images containing field or tennis court. Each category is a combination of the original categories in parentheses. These filtered images are randomly divided into training set, validation set, and test set, with the proportions being 80%, 10%, and 10%, respectively.

The Potsdam dataset is divided into six most common land cover classes, namely impervious surfaces, Buildings, Low vegetation, Trees, Cars, and Clutter(background). Clutter includes water, containers, tennis courts, swimming pools, etc. The dataset contains 38 aerial orthographic images of urban areas with a size of 6000x6000, and the ground resolution is 0.05m. Due to the limitation of GPU memory, multiple 512x512 images are cropped from each image. These cropped images are randomly divided into training set, validation set, and test set, with the proportions being 60%, 20%, and 20%, respectively.

### 4.2 .Experimental setup and evaluation metrics

In this paper, U-Net is adopted as the backbone of DSSN. In the optimization phase, we use the cross entropy loss function and Adam backward propagation optimization algorithm [42] with the learning rate 10e-4. The superpixel segmentation method uses Simple Linear Iterative Clustering (SLIC) [43]. The number of superpixel segmentation $K$ and the confidence threshold $F_t$ are set to 1000 and 0.7, respectively.

The evaluation of classification uses Overall Accuracy (OA) and Mean Intersection over Union (mIOU), defined as Eq. (8) and Eq. (9).

$$OA = \frac{TP+TN}{TP+FP+TN+FN} \quad (8)$$

$$mIOU = \frac{1}{n}\sum_{1}^{n}\frac{TP}{TP+TN+FN} \quad (9)$$

where $TP$ is the number of pixels with positive classes which are correctly classified. $TN$ is the number of pixels with negative classes, which are correctly classified. $FP$ is the number of pixels with positive classes, which are misclassified. $FN$ is the number of pixels with negative classes, which are misclassified. $n$ is the number of classes.

**4.3 Sensitivity analysis of critical parameters**

This study forms a closed loop consisting of a DSSN and an ontology reasoner, which iterates continuously to improve the accuracy of classification. As shown in Fig. 3, along with the increase of the number of iterations, the accuracy of classification will continue to improve. When it reaches a certain level, the accuracy will stop improving or even decrease. The optimal number of iteration on the UCM dataset is 3, and the performance of the Stage II is about 3% higher than that of the Stage I. The optimal number of iteration on the Potsdam dataset is 4, and the performance of the Stage II is about 2% higher than that of the Stage I.

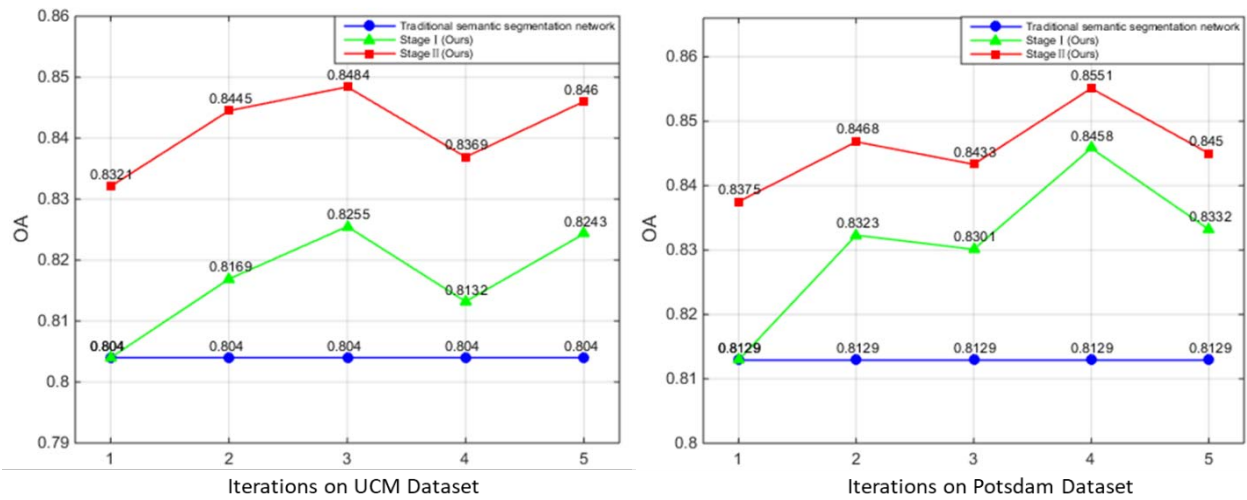

Fig. 3. The performance variation of the presented method under different iterations.

The classification results of this method on the UCM dataset and the estimation of shadow and elevation are shown in Fig. 4. Fig. 5 shows the classification results on the Potsdam dataset and the estimation of shadow and elevation. In the figure, the first column is the input image, the second column is the ground truth, the third column is the classification of U-Net, and the fourth and sixth columns are the outputs of Stage I in Iteration 1 and Iteration 3, respectively. The 5th and 7th columns show the outputs of Stage II in Iteration 1 and Iteration 3, and the 8th-11th columns show the estimation of shadow and relative elevation. It can be seen from Fig. 4-5 that the boundary of the classification (Stage I) is more accurate, which indicates that the DSSN improves the accuracy of classification by using extra information obtained by ontological reasoning. From the Stage I to the Stage II, our method directly corrects some misclassifications and the results of correction are interpretable. As shown in Fig. 4, the last image in the third column, the area misclassified as vegetation is surrounded by ground. This phenomenon is unlikely to occur in this scenario, so the reasoner in the Stage II classifies the area as ground according to the rules of ontological reasoning. For the same reason, the area misclassified as ground in the first picture of the third column is judged as road.

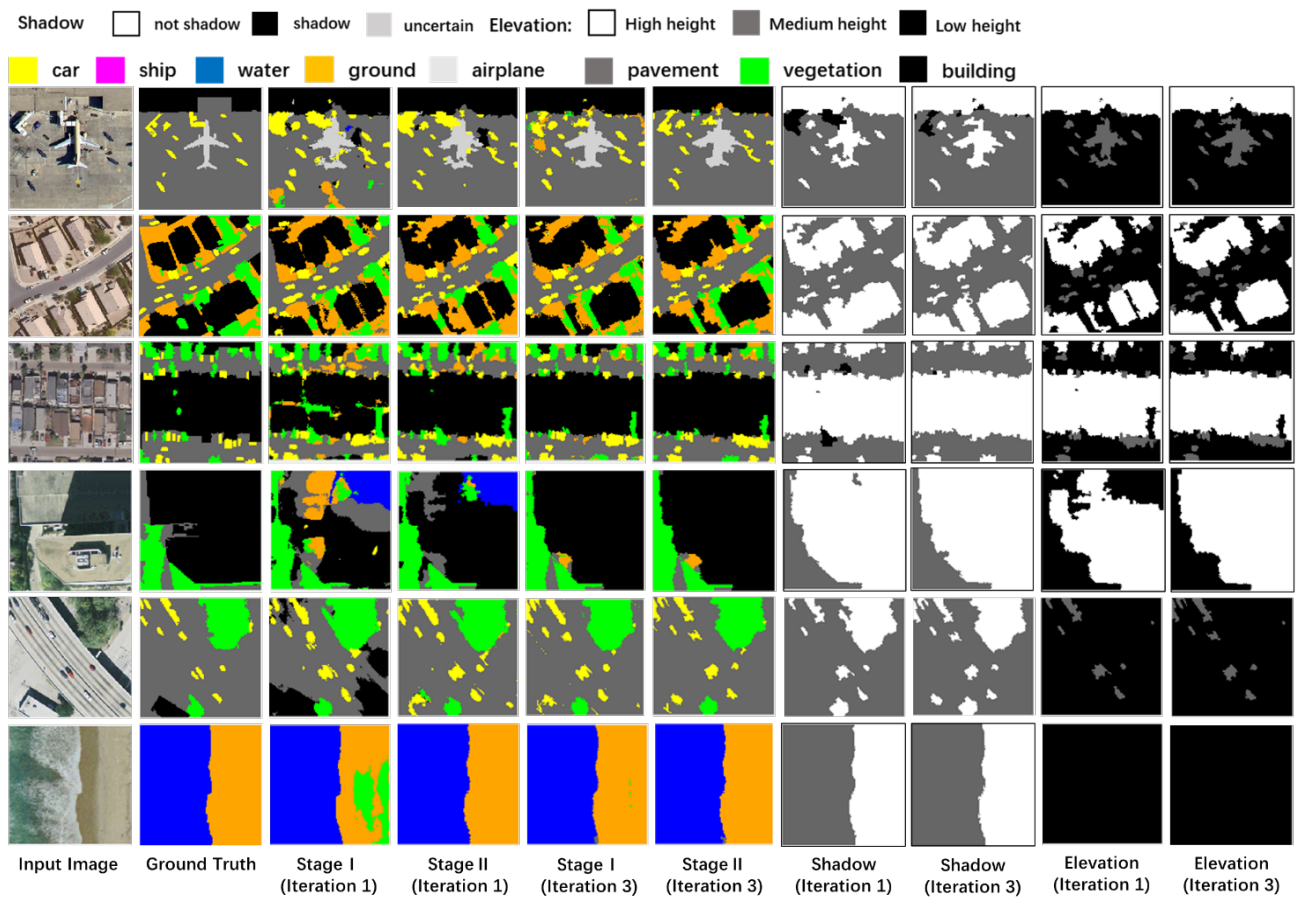

Fig. 4. The classification result and the prediction result of shadows and elevations on the UCM dataset.

It can be seen from Fig. 4 and Fig. 5 that the estimates of shadow and relative elevation are accurate, for the contours of the objects in the images are basically outlined. As the number of iterations increases, the classification result becomes more accurate with the shadow and relative elevation being more accurate. The extra information extracted from the classification is also used as additional input of the classifier to assist classification. The classifier and the reasoner interact with each other and promote together.

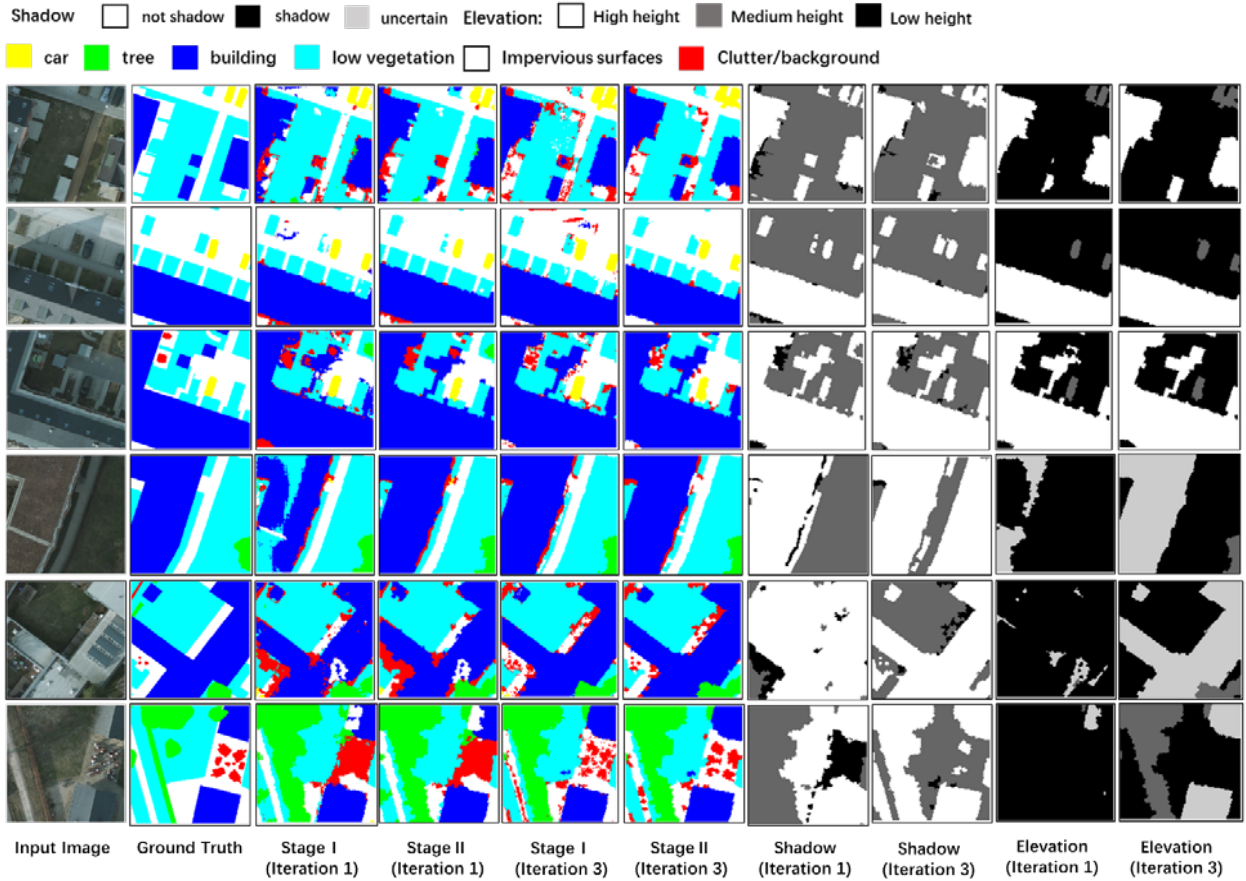

Fig. 5. The classification result and the prediction result of shadows and elevations on the Potsdam dataset.

## 4.4 Comparison with the state-of-the-art methods

To fairly show the effectiveness of our proposed CBF, we compare our presented CBF with two baselines on UCM and Potsdam. The baselines include U-Net [37] and Semantic Referee [36]. It can be seen from Fig. 6 and Fig.7 that the classification of the Stage I and the Stage II of our CBF are better than the U-Net and the Semantic Referee. The classification performance of Stage II is better than Stage I.

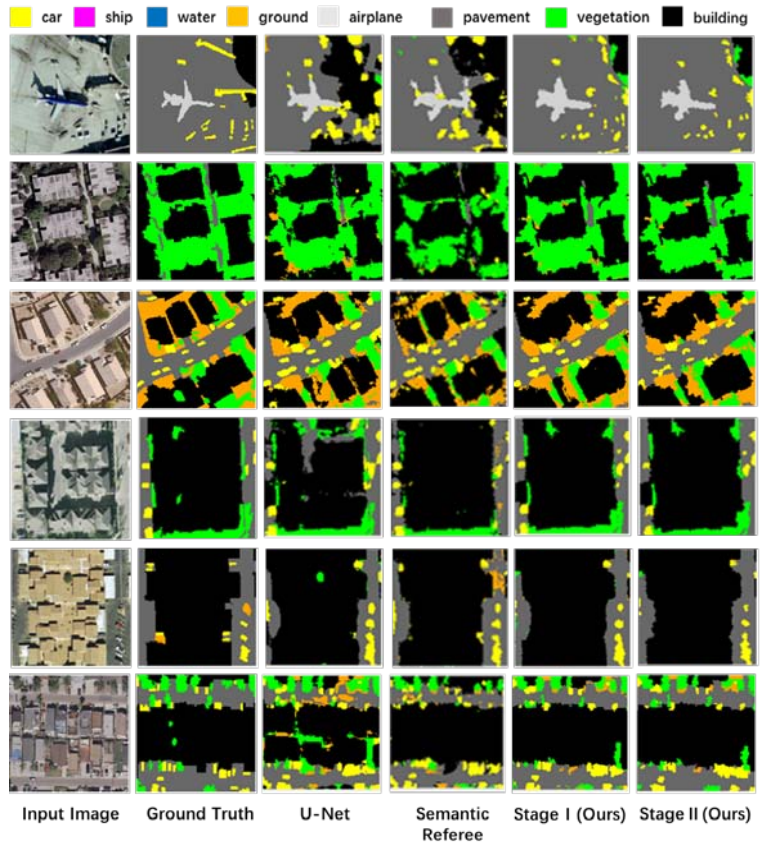

Fig. 6. The visual classification results of our proposed method and other baselines on the UCM dataset.

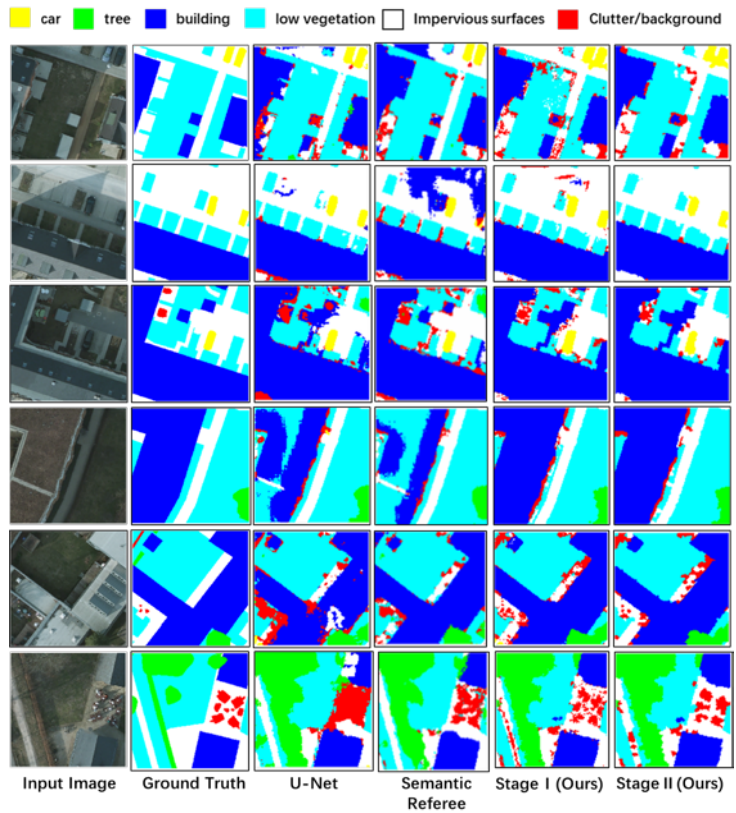

Fig. 7. The visual classification results of our proposed method and other baselines on the Potsdam dataset.

As shown in Table 3, the classification performance of different methods on the UCM dataset is summarized. The classification performance of different methods on the Potsdam dataset is shown in Table 4. It can be seen from Table 3 and Table 4 that all of the methods which combine deep learning and ontological reasoning (i.e., Semantic Referee and our proposed CBF) outperform the DSSN alone (i.e., U-Net).

Table 3. The classification performance of different methods on the UCM dataset under OA and mIOU

| Method | OA | mIOU |
| --- | --- | --- |
| U-Net [37] | 0.8026 | 0.6606 |
| Semantic Referee [36] | 0.8228 | 0.6780 |
| Stage I（Ours） | 0.8374 | 0.6885 |
| Stage II（Ours） | **0.8592** | **0.7098** |

Table 4. The classification performance of different methods on the Potsdam dataset under OA and mIOU.

| Method | OA | mIOU |
| --- | --- | --- |
| U-Net [37] | 0.8129 | 0.6444 |
| Semantic Referee [36] | 0.8276 | 0.6669 |
| Stage I（Ours） | 0.8458 | 0.6764 |
| Stage II（Ours） | **0.8551** | **0.6893** |

Among all the methods in Table 3 and Table 4, our proposed CBF achieves the highest OA and mIOU, which fully verified the effectiveness of our proposed method. Compared with the Stage I, the classification performance of the Stage II, which corrects misclassification by intra-taxonomy ontological reasoning, is significantly improved. It proves that the ontological reasoning helps to improve the classification performance. Deeply, Stage I can outperform the traditional DSSN (i.e., U-Net) shows that the additional information, which is referenced by the extra-taxonomy ontological reasoning module, benefits lifting the discriminative performance of DSSN from the input perspective.

## 5. Conclusion

In this paper, we present the new CBF to collaboratively combine data-driven deep learning and knowledge-guided ontological reasoning where domain expert knowledge is expressed by ontology to effectively mine rich semantic information in the RS imagery. It is worth noting that CBF is a general idea and more variants can be designed based on the specific task requirements. In this paper, ontological reasoning includes intra-taxonomy ontological reasoning and extra-taxonomy ontological reasoning. Both the intra-taxonomy ontological reasoning directly corrects misclassification and the extra-taxonomy ontological reasoning provides additional information to assist deep learning indirectly, which not only further improve the accuracy of classification based on deep learning methods, but also enhance interpretability by knowledge reasoning. Extensive experiments on two publicly open RS datasets such as UCM and ISPRS Potsdam shows that our proposed CBF can obviously outperform the traditional DSSN (e.g., U-Net) and existing ontological reasoning based method (e.g., Semantic Referee).

To further improve the ontological reasoning performance, more types of ontological reasoning rules such as boundary constraints should be constructed. The shadow and elevation can effectively improve the classification accuracy as the input of additional channel of the DSSN, so it is necessary to further explore the auxiliary mechanism of the additional information to the DSSN. Objectively, the proposed joint method in this paper only interacts with the DSSN and ontological reasoning from the input and output perspectives. In our future work, we will try to integrate deep learning and knowledge reasoning at a deeper level to achieve autonomous learning in the reasoning.

**Acknowledgments**

This work was supported by the National Key Research and Development Program of China under grant 2018YFB0505003; the National Natural Science Foundation of China under grant 41971284; the China Postdoctoral Science Foundation under grant 2016M590716 and 2017T100581; the Hubei Provincial Natural Science Foundation of China under grant 2018CFB501.